\documentclass{article}

\usepackage{arxiv}

\usepackage[utf8]{inputenc} % allow utf-8 input
\usepackage[T1]{fontenc}    % use 8-bit T1 fonts
\usepackage{hyperref}       % hyperlinks
\usepackage{url}            % simple URL typesetting
\usepackage{booktabs}       % professional-quality tables
\usepackage{amsfonts}       % blackboard math symbols
\usepackage{amsmath}        % math environments and symbols
\usepackage{nicefrac}       % compact symbols for 1/2, etc.
\usepackage{microtype}      % microtypography
\usepackage{lipsum}		% Can be removed after putting your text content
\usepackage{graphicx}
\usepackage{natbib}
\usepackage{doi}

\usepackage{float}
\usepackage{subfig}
\graphicspath{{images/}}
\usepackage{hyperref}
\usepackage{multirow} %for tables

\usepackage{float}
\usepackage[table]{xcolor} % Allows row coloring
\usepackage{array}
\usepackage[acronym]{glossaries}
\usepackage{xcolor}

\usepackage{subcaption} % for subfigure environment

\title{Development of an Intuitive GUI for Non-Expert Teleoperation of Humanoid Robots}

\date{September 13, 2025}	% Here you can change the date presented in the paper title
\author{ 
    Austin Barrett\\
	School of Engineering and Computer Science\\
	Laurentian University\\
	\texttt{abarrett1@laurentian.ca} \\
	\And
    Meng Cheng Lau \\
	School of Engineering and Computer Science\\
	Laurentian University\\
	\texttt{mclau@laurentian.ca} \\
    \href{https://orcid.org/0000-0003-3517-4900}{\textbf{https://orcid.org/0000-0003-3517-4900}}\\
}

\renewcommand{\shorttitle}{}

\hypersetup{
pdftitle={Development of an Intuitive GUI for Non-Expert Teleoperation of Humanoid Robots},
pdfsubject={cs.RO, cs.LG},
pdfauthor={Austin Barret, Meng Cheng Lau},
pdfkeywords={},
}

\begin{document}
\maketitle

\begin{abstract}
The operation of humanoid robotics is an essential field of research with many practical and competitive applications. Many of these systems, however, do not invest heavily in developing a non-expert-centered graphical user interface (GUI) for operation. The focus of this research is to develop a scalable GUI that is tailored to be simple and intuitive so non-expert operators can control the robot through a FIRA-regulated obstacle course. Using common practices from user interface development (UI) and understanding concepts described in human-robot interaction (HRI) and other related concepts, we will develop a new interface with the goal of a non-expert teleoperation system.
\end{abstract}

\begin{center}
\href{https://sites.google.com/laurentian.ca/limrl}{\textbf{Laurentian Intelligent Mobile Robotic Lab}}
\end{center}

\keywords{HuroCup  \and User Friendly \and ROS \and GUI}

% keywords can be removed
% \keywords{Humanoid locomotion \and Reinforcement learning \and Reduced-order model \and Motion imitation \and Adversarial learning \and Gait generation}

\section{Introduction}

Teleoperation in humanoid robotics has grown rapidly over the past decade, driven by advances in robotic movement and control. These systems are essential in domains like healthcare, mining, and space exploration \cite{liu2018remote, godelius2021remote, gomez2021robotics}. Competitions such as the FIRA RoboWorld Cup—and specifically its HuroCup teleoperation event—serve as testing grounds for teleoperated humanoid systems \cite{DRCrulesdoc}. A core challenge in this event is navigating an obstacle course without contact, which demands a clear and intuitive user interface. Designing such a GUI requires strong visual feedback and simplified control schemes, making the system accessible to users with no prior experience. This is especially important in team settings like FIRA, where new participants frequently inherit systems built by others. A usable interface not only enhances operator performance but also reduces the need for retraining and lowers cognitive workload \cite{white2020design, rea2022still}. This research aims to create a user-friendly GUI tailored for the FIRA HuroCup obstacle run. It will prioritize camera-based visualization, clear communication of robot state, and scalable design for broader applications. The GUI will be evaluated qualitatively against a previous version using standard competition-style obstacle courses, though testing with non-experts and quantitative metrics remains outside the current scope due to time constraints \cite{zhang2023teleoperated, sankar2023systematic}.

% [Sections from Hardware to Simulation remain unchanged unless you want to specify any updates for Kid Size hardware/software]

\section{Literature Review}

%Teleoperated humanoid robots are advancing rapidly, spurred by events such as the FIRA RoboWorld Cup, yet most interfaces still assume expert operators. Research stresses that GUIs must let novices control robots quickly and safely, broadening access to the technology
Prior research on teleoperated humanoid robots has emphasized the importance of user interface design, particularly in the context of competitions like the FIRA RoboWorld Cup. Despite ongoing advancements, many systems still rely on expert operators. Studies underline the need for GUIs that support fast and safe control by novice users, thereby improving accessibility~\cite{rea2022still}. The current interface, lacking clear labels, intuitive layouts, key actions like crawl, coherent camera placement, and on-the-fly motion tuning, illustrates these shortcomings. Teleoperation bridges human intent and robot action \cite{darvish2023teleoperation}. Effective systems build situational awareness, manage operator workload, and mitigate error \cite{steinfeld2006common}. Shared-control schemes further ease cognitive load by blending autonomous and manual control \cite{adamides2014human}. Intuitive visual feedback and adaptive GUIs cut training time for non-experts~\cite{goodrich2013teleoperation}, while task-specific visuals condense complex manoeuvres into simple commands \cite{sankar2023systematic}. Scalable architectures, such as the DRC-HUBO interface, show how one design can span competitions and industry \cite{zucker2015general}. Performance is typically evaluated through error rates and user-satisfaction metrics \cite{steinfeld2006common}.

Classic usability texts reinforce these ideas: maintain consistent navigation and colour schemes, hide rarely used controls, provide immediate feedback, and adapt fluidly to different screens~\cite{Beaird2020, Boss2016, Karray2013}. Applying such rules to teleoperation reduces novice frustration and errors. Evidence from high-risk domains underscores the value of friendly teleoperation. In mining, remote control shields workers from danger \cite{godelius2021remote}. Surgical robots hinge on intuitive human–robot interaction to ensure safety \cite{sun2023overview}. Competitions amplify these needs: DARPA Robotics Challenge teams navigated disaster scenarios with interfaces balancing shared autonomy and operator clarity \cite{Marion2018}, and FIRA rules similarly demand fast, reliable teleop \cite{DRCrulesdoc}. Iterative, user-centred design (UCD) drives better GUIs~\cite{mao2005state} and high cognitive load invites confusion and errors \cite{goodrich2007human, Chen2007}, so refinements that lighten mental effort—such as cleaner layouts or automated feedback loops—improve speed and accuracy \cite{tagliamonte2024generalizable}. Although System Usability Scale (SUS) questionnaires and similar tools can quantify these gains, formal user trials may not always be feasible \cite{steinfeld2006common}. Despite progress, most teleoperation research still targets trained specialists. Interfaces explored in joystick, VR, or wearable studies prioritise raw capability over novice accessibility \cite{Franz2013, ryu2004multi, Fritsche2015}. Few offer GUIs that are simultaneously simple, extensible, and competition-ready. This project, therefore, proposes a redesigned interface that merges HRI guidelines, web-design best practices, and shared-control paradigms to enable non-expert teams to pilot humanoid robots effectively on the FIRA obstacle course and beyond.

\section{Methodology}

This research aims to develop a user-friendly GUI for non-expert operators to control humanoid robots in the FIRA HuroCup teleoperated obstacle run. The GUI addresses a gap in existing research by prioritizing usability and intuitive design for users with no prior experience in teleoperation systems \cite{rea2022still}. Relying primarily on camera input, the interface emphasizes effective visualization of the robot’s environment and operations. Development is guided by an iterative cycle:
\begin{itemize}
\item Create wireframes and identify key features per iteration.
\item Implement and adapt features throughout development.
\item Test feature usability and assess layout comfort.
\item Refine based on observations before the next cycle.
\end{itemize}

This structured, iterative methodology ensures consistent improvement across interface versions, enhancing usability, layout, and functionality.

%\subsection{Research Design}

%Interface development follows multiple design iterations to ensure that all added features serve a functional purpose without disrupting overall usability. This approach—common in UI and web development—promotes consistency and refinement \cite{Wynn2017}. While prior versions of the GUI used Bootstrap and other pre-built elements, this project emphasizes custom development with HTML, CSS, and JavaScript for greater control and adaptability.

\subsection{GUI Platform and Tools}

Interface development typically involves multiple design iterations to ensure that each added feature enhances functionality without compromising overall usability. This iterative approach, widely used in UI and web development, supports consistency and continuous refinement \cite{Wynn2017}. While earlier versions of the GUI relied on Bootstrap and other pre-built components, the current project focuses on custom development using HTML, CSS, and JavaScript to achieve greater control and flexibility.

The GUI is built for a bipedal mobile robot classified under FIRA’s kid-size humanoid standards \cite{jacky2025}. Inputs are limited to the camera and terminal console, with no external sensors available. This limitation highlights the importance of optimizing camera-based environmental awareness for the operator. The system runs on Ubuntu Mint 16.07 with ROS1, using Python 2.7 and C++ in a publisher-subscriber model. A ROS node for teleoperation exists, developed by a past student, and provides a basic browser-based GUI with limited functionality. Existing buttons allow for basic motion, while features like crawling or lateral movement are either absent or inconsistently implemented. Some elements, such as an obstacle mapping widget, are partially functional but incomplete. The interface communicates through a server-based node, with front-end development mimicking a standard web application using HTML, CSS, and JavaScript.

\subsection{Development and Testing Procedures}

%Development began by implementing critical controls—movement and camera views—and ensuring they are intuitive and responsive. Additional components include crawling, voltage indicators, and logs for diagnostics. With FIRA disallowing additional sensors, all feedback must come from internal data and camera visuals. Wireframes guided the placement of primary features, which were refined as development progressed. Interactivity is managed with JavaScript, while styling and layout rely on HTML and CSS. No external libraries are used to maintain granular control and ease the iterative process.

Development began with the implementation of essential controls, such as movement and camera views, ensuring they were both intuitive and responsive. Additional features were added, including crawling functionality, voltage indicators, and diagnostic logs. Due to FIRA's restriction on external sensors, all feedback is derived solely from internal data and camera visuals. Wireframes were used to guide the layout of core elements, which were continuously refined throughout the development process. Interactivity is handled using JavaScript, while HTML and CSS define the layout and styling. To maintain full control and support iterative improvements, no external libraries were used.

Communication with the robot uses HTTP for commands and WebSockets for real-time feedback, with JSON used for message formatting. Testing was performed on a regulation-compliant turf track with obstacles to simulate competition conditions. Usability evaluations were based on task success and user workflow efficiency. Features were retained or revised based on their contribution to task performance. Iterative development enables continuous improvements and refinements, supporting a more user-friendly and task-oriented interface over time.

\subsection{Ethical and Practical Constraints}

Due to time constraints, no external user testing or ethics approval was pursued, and evaluation was conducted solely by the developer. Future work could incorporate third-party evaluations and quantitative testing. Additionally, inconsistent robot behavior, caused by modifications from other users, sometimes affected testing. These irregularities, while not part of the GUI design, were accounted for during evaluations by focusing on ideal performance conditions.

\section{Implementation}

\begin{figure}[H]
    \centering
    \includegraphics[width=0.6\textwidth]{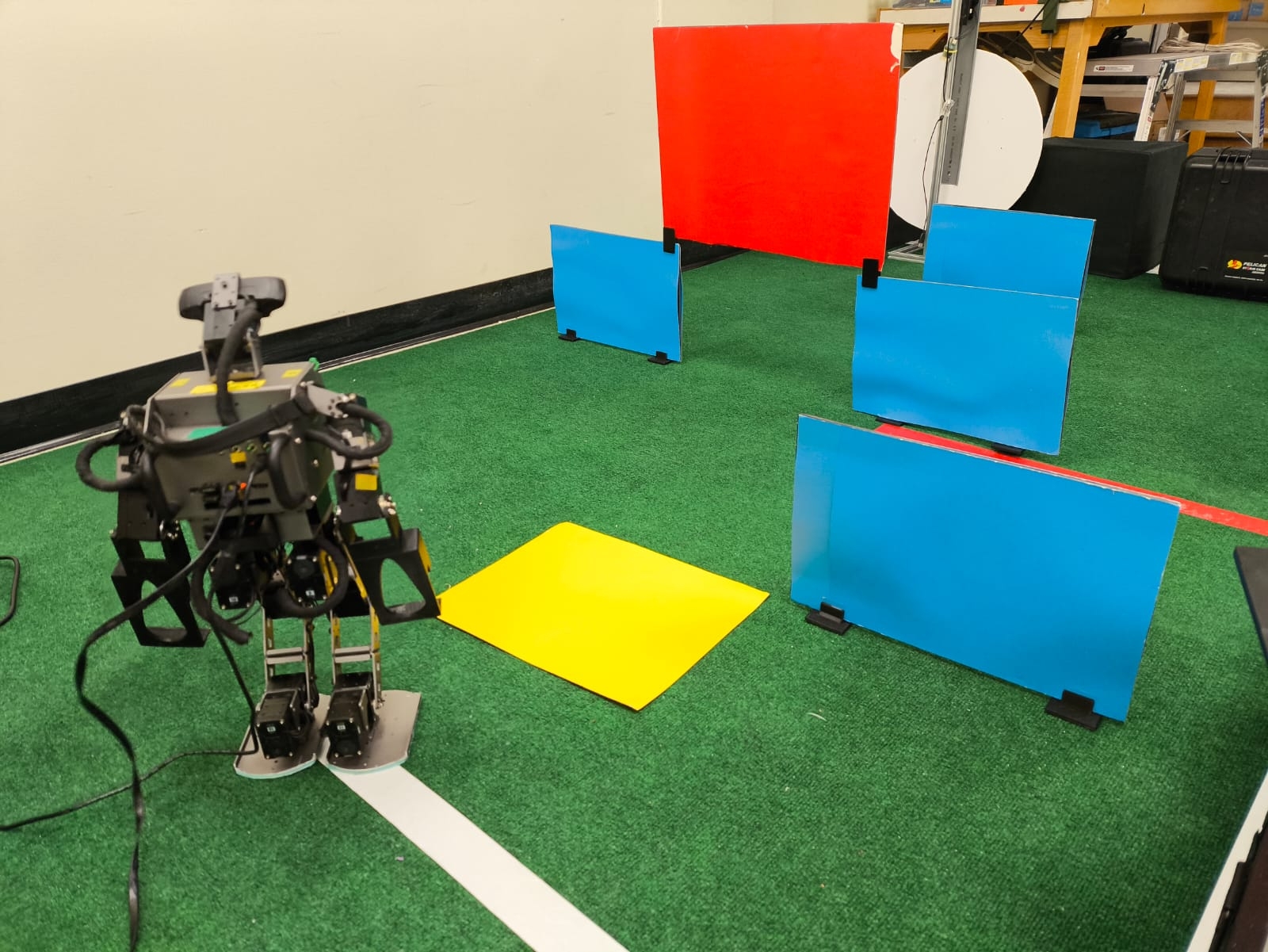}
    \caption{Visual of kid-size humanoid robot navigating a replica of the FIRA obstacle run event.}
    \label{fig:event}
\end{figure}

The main goal of the implementations is to develop a user-friendly interface for teleoperation in the FIRA RoboWorld Cup. Focusing on making it as easy for operators to navigate and utilize the robot for its intended use in the obstacle run, as shown in Fig.~\ref{fig:event}.

The implementation will closely follow the design philosophy and concepts that were previously outlined in the methodology chapter. Such as focusing on good user interface development tactics, an iterative development approach that uses testing in realistic application environments, and a focus on the camera as the main user feedback for robotic movements and environmental situations.

\subsection{System Architecture and Communication}

The development of the software to complete the task is highly dependent on the operating systems and software the robot already utilizes. In this case, the robot uses a distribution of Ubuntu mini with the ROS (Robot Operating System) installed for the operation of the implemented code. The operating system specifically utilizes version Mint 16.04 of the Ubuntu operating system, which allows for a lightweight operating system to perform actions on. As for the ROS software, it is what is mostly worked with during the development of the UI, where ROS is the middleware that utilizes a node-based structure for the execution of robotic actions. This means that development involves updating features of the remote-control node system.

The node system utilizes Python 2.7 for the development of the code. In this specific application, we activate the remote server node, which runs the communication for the actions the users perform in the interface to the actual robotic actions that they correlate to. This node structure sets up the local server, which can be accessed via the IP 10.42.0.1 in the browser. At this point, the front end created by the node is visible, utilizing HTML/CSS and JavaScript for user interactions. Allowing for the development of the UI to be similar to the web development structure in terms of coding and conventions.

As for possible JavaScript library utilization, the use of jQuery for certain UI interactions, and mjpegcanvas.js for streaming video is utilized. \textit{mjpegcanvas} allows users to utilize the \textit{img} tag and output the live camera feed to the element for the user to view and interact with.  
\textit{Jquery} is used within the JavaScript for DOM manipulation, handling button presses, and hiding or showing UI elements such as submenus. Its purpose is to reduce the amount of boilerplate code used to select and modify elements of the page.

The robot utilizes ROSbridge to bridge communication between ROS and the web-based GUI. ROSbridge runs on the robot as a WebSocket server, exposing ROS topics and services in JSON format. On the client side, \textit{roslib.js} is used to establish a WebSocket connection with ROSBridge, allowing the web interface to publish movement commands and subscribe to feedback topics. This setup removes the complexity of ROS network protocols and enables real-time control and monitoring from a standard browser without requiring native ROS installations.

For this web-based node, it utilizes both rosbridge and \textit{roslib.js}. ROSbridge is what allows the node to create the local server (websocket server), and \textit{roslib.js}, included in the web page, opens a WebSocket connection to Rosbridge. With both of these utilities, we can send ROS messages over the WebSocket to have effective communication with the user and the robot. This setup abstracts away the complexity of handling ROS network protocols (like \textit{roscpp} or \textit{rospy}), making it possible to control and monitor the robot from a standard web page.

The process in which the user communicates with the robot follows the path of the user first interacting with the robot. In this case, it would be the UI, and then that action is converted into a readable JSON for ROS by \textit{roslib.js} and sent over the Rosbridge WebSocket. At this point, that command is read and then performed as an action by the robot. These possible messages will contain all the needed actions to perform the obstacle run event successfully and will require the user to utilize the proper commands given a situation.

The Messages contain the action that is to be performed, which is then sent to the remote control node that is subscribed to the actions. The node reads each incoming message and interprets it as a request to walk (forward/backward), turn, or shift the robot’s body.

\subsection{GUI development process}

\begin{figure}[htp]
    \centering
    \includegraphics[width=0.9\textwidth]{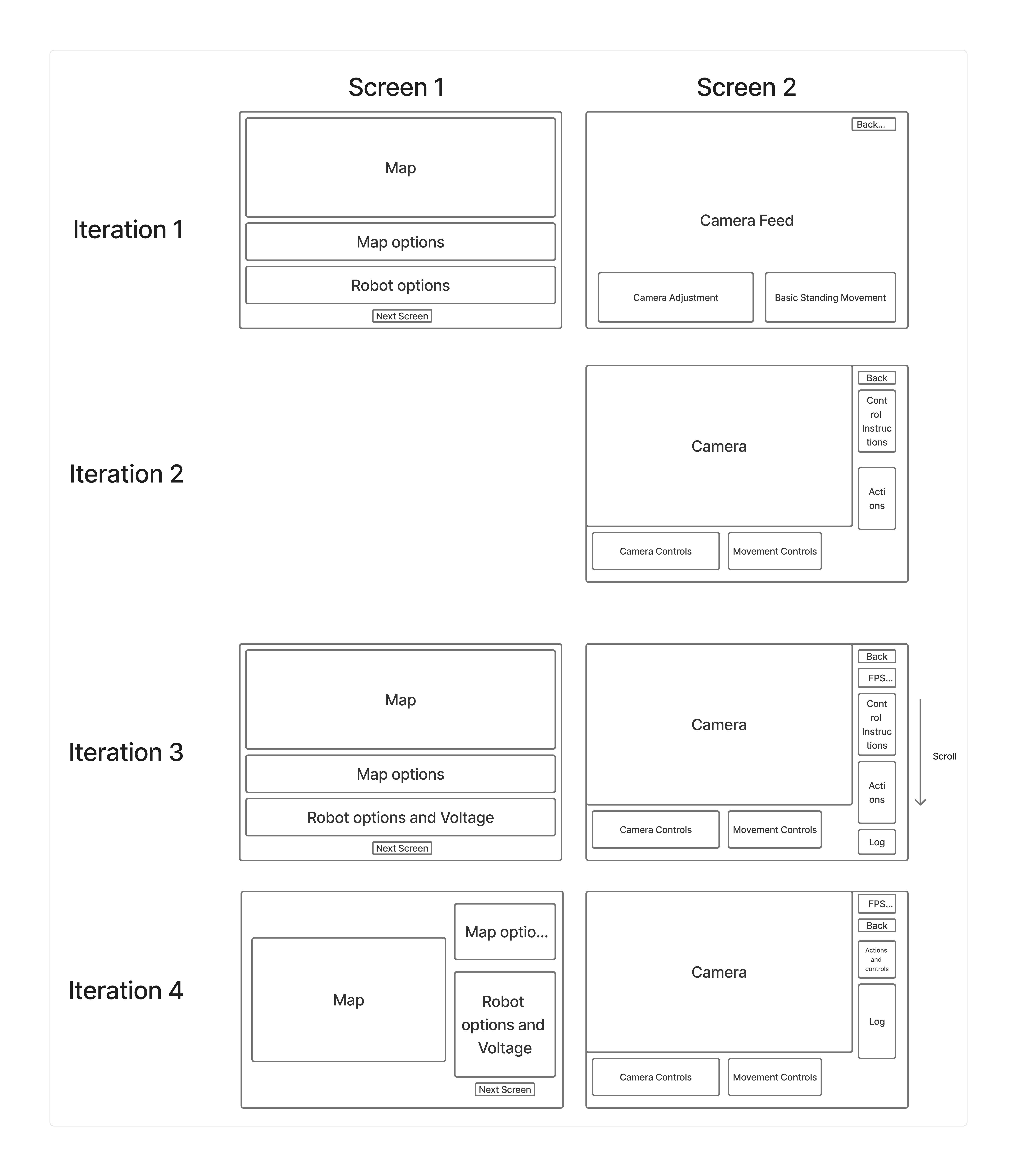}
    \caption{Wireframe depiction set before each draft}
    \label{fig:gui_wireframe}
\end{figure}

Due to the utilization of an iterative process during development, the first version of the GUI draft contained the goals of displaying content in a non-cluttered way to the user and making sure the very basic movement (not everything needed to complete the obstacle course) was possible, and newer features were to be added when the first development was done. It is also important to note that a basic version of this already existed in the robot, but it contained critical functionality flaws, unintuitive design, and missing features. There was a feature that was partially implemented that allows the user to create a map that shows the robot's location along with the location of some obstacles in a rudimentary fashion, which is elaborated upon shortly. Also note that for each iteration of this development process, A rough draft of a wireframe model was created to allow the development process to follow some structure as features were implemented, as shown in Fig.~\ref{fig:gui_wireframe}.

The first draft allowed us to see what was possible with the UI and utilized the simple technique of putting similar operations in the same area on the screen. This is known as proximity or grouping \cite{Lidwell2010}. In this case, the application of grouping involved putting the controls for basic movement on screen together, along with making 2 screens for the setup of the map and the actual control environment of the robot. This process is continually utilized for a cohesive interface throughout the rest of the development. Another important feature during this iteration is the new placement of the camera at the forefront of the user's attention, this was done as it is the most important interaction between the human and robot in terms of communication and the fact that in the previous system, it was not scaled to a perspective that was easy for the user to preserve without difficulty in some scenarios. The final thing for this iteration to focus on would be to implement the basic shifting movements for left and right movement, which were not present in the old system. Adding these features will complete the umbrella of basic movement and provide the user with intuitive movement to navigate the obstacle course. 

This first iteration provided good results and showed that the placement of the movement controls for the robot and the addition of the left and right shift movements proved as a great success in the tests, working as expected. Through testing, one point of concern was the camera size. As the camera increases in size on the screen, the throughput decreases, and depending on the state, it becomes detrimental to operation. So future iterations would scale down the camera and utilize the additional space to focus emphasis on controls to balance the robot's vision with controls. With this, the second iteration of implementation was set.

The preceding iteration involved mapping the controls to common keyboard controls, so that, along with pressing the buttons on the screen, the keyboard can be utilized. This requires adding all the actions needed and mapping them to logical keys. Due to the need for the movement of the camera along with the movement of the robot, the utilization of 2 directional inputs via keyboard where needed so due to the convention in the modern moment in keyboards being attributed to the W, A, S, and D buttons, those where selected to be used for the robot movement and the arrow keys for camera movement \cite{Krug2005}. Alternative actions also needed to be added, such as the get-up and crawl actions. This, along with resizing the camera for optimal size/throughput rate for operations, was the main goal of this iteration. Also, due to the notes in the last chapter based on the camera, we would move the camera to have its own section of the screen without the controls being overlaid on top of it. This would reduce the camera resolution and improve the throughput of the live image over the network. There would also be the translation of basic control instructions for new users that were present in the previous system. 

For this iteration, the camera was demonstrated to be much better in performance, and the binding of the movement buttons proved to be beneficial in simplifying the movement controls. During testing, there was an exception made for this button binding, and that was the action buttons, as they were added as selectable buttons on the screen rather than being bound to keys. This was done for a few reasons. The first is that they are less commonly used in the application of obstacle runs than the other movements; thus, their priority is less prudent. With this priority in mind, their prevalence in screen real estate is also much smaller than the other controls. This follows the design approach in UI development known as visual hierarchy, which gives the distinction of more important things to be more prevalent on the screen for the user \cite{Johnson2010}. And my second reason is due to the sub-goal of making this project scalable. If these actions are mapped to a key bind, other actions are later added and mapped. Eventually, it can become confusing to the user what button completes what actions, and could lead to operator overload and human error in actions \cite{Goodrich2013}. So with this in mind, the actions are simply left as selectable buttons. And with this iteration, we have a UI that can complete the obstacle course, but can be improved upon.

Another refinement added during the third iteration was the inclusion of start and reset buttons on the right-hand sides. These buttons set the position of the robot in their respective manners. The start button puts the robot in the default starting position for the event, and the reset puts the robot in a default standing position. This is useful if the operator is getting confused as to what the current positioning of the robot is; they can simply reset the position so that they can continue the operation from a point they already understand.

The third iteration for development would look into adding some of the internal senses to the UI or moving them so that the user could access these during operation. This would allow the user to gain more information about the status of their robot and possibly troubleshoot the problem on their own. These include the rendering frames per second (FPS), battery life, and log of backbend messages. The battery voltage indicator and log were both present in the old system, and just need to be reformatted to fit the new layout. 

After developing these features, minimal testing was needed to see their application validity. They did not hinder current operations and provided useful information on demand, so keeping these features proved to be beneficial.

The fourth iteration had the goal of improving the visual experience of all the components within the GUI. So some of the methods implemented to improve this were High-Contrast Visuals, which allow the controls to be more visible and prudent compared to the general camera feed. Along with utilizing colours to correlate different portions of the screen. Dynamic screen sizing was also important for this phase a,s the previous iterations had been developed on a full 1080p display but in competition applications that will likely be a much smaller window to conserve as much processing power as possible and to fit on the smaller laptop screens, so dynamic sizing web principles are important. Now, the placement and labelling of certain features on the display also become more prominent in development. The addition of collapsible menus can also allow users to only open the menus needed for operations. For example, the control instruction menu can be collapsed so that returning users can minimize the menu so they don't need to have the menu take up space that can be otherwise used for controls.

First, the blue and orange were selected as contrasting colours so that different features could be associated with colour. In the case of this application, blue and orange were selected as the colours as they are in most applications, visible when compared to the screen live. The buttons for controls are located in bright orange on the screen and are larger and brighter than their surrounding components, as they are more important for the operation of the telescope. Labelling was also elaborated and refined with simple descriptions and a pop-up instruction option so users could choose to have it on screen.

Additionally, on the final iteration of the development cycle, a new system on the starting screen was created, where the user can make micro-adjustments to the duration of the walking movements on the fly. Because the system uses JS listeners, and individual button presses without holding the button produce one movement. So if you're on an obstacle course and need to only move forward a unit smaller than any derivative the the default value, you can just lower the value to be smaller so the user can make the desired distance move without affecting the result. This feature utilizes a subscriber-listener model where JavaScript event listeners capture the user’s adjustments to walking movement coefficients in real time. Specifically, when a user modifies the distance or angle values in the input fields on the GUI, a listener function stores these updated values locally in the browser’s session storage. Upon pressing a movement button, another listener retrieves the current coefficients and constructs a movement command message. This message is then published over a WebSocket connection using roslib.js to the robot's ROSbridge server. On the robot's side, a ROS node subscribes to the movement topic, reads the received message, and then executes the movement using the user-defined values. This model allows seamless, on-the-fly customization of robotic movements, allowing the user to fine-tune actions without restarting the system.

\subsection{Refinements and Iterations}

During the iterative process of development, there were many refinements to the old code and the newly written code. The following depicts the changes of some of these features and fine tuning that was completed during the development period.

\begin{figure}[htp]
    \centering
    \includegraphics[width=0.4\textwidth]{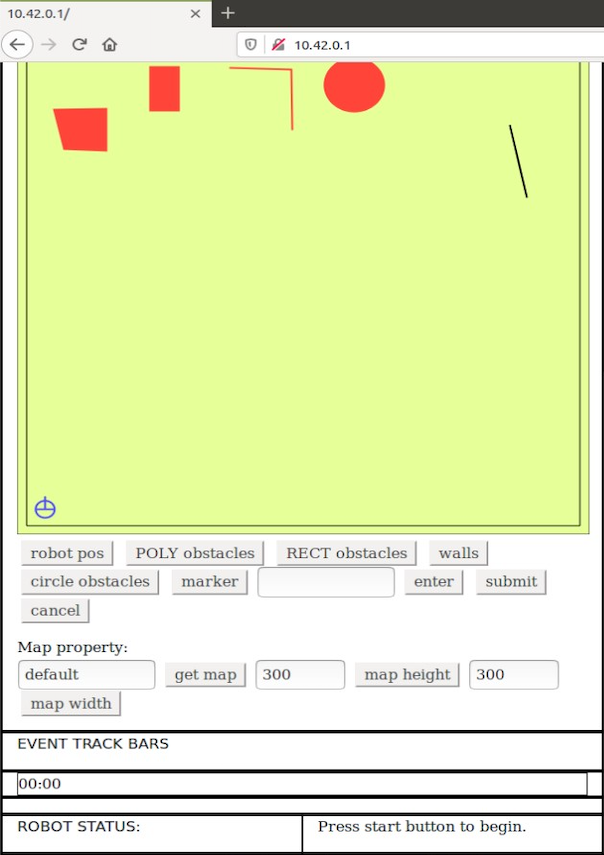}
    \caption{Image of original screen with the map on the left side of the screen and operation buttons on the bottom}
    \label{fig:old_map_image}
\end{figure}

\begin{figure}[htp]
    \centering
    \includegraphics[width=1.0\textwidth]{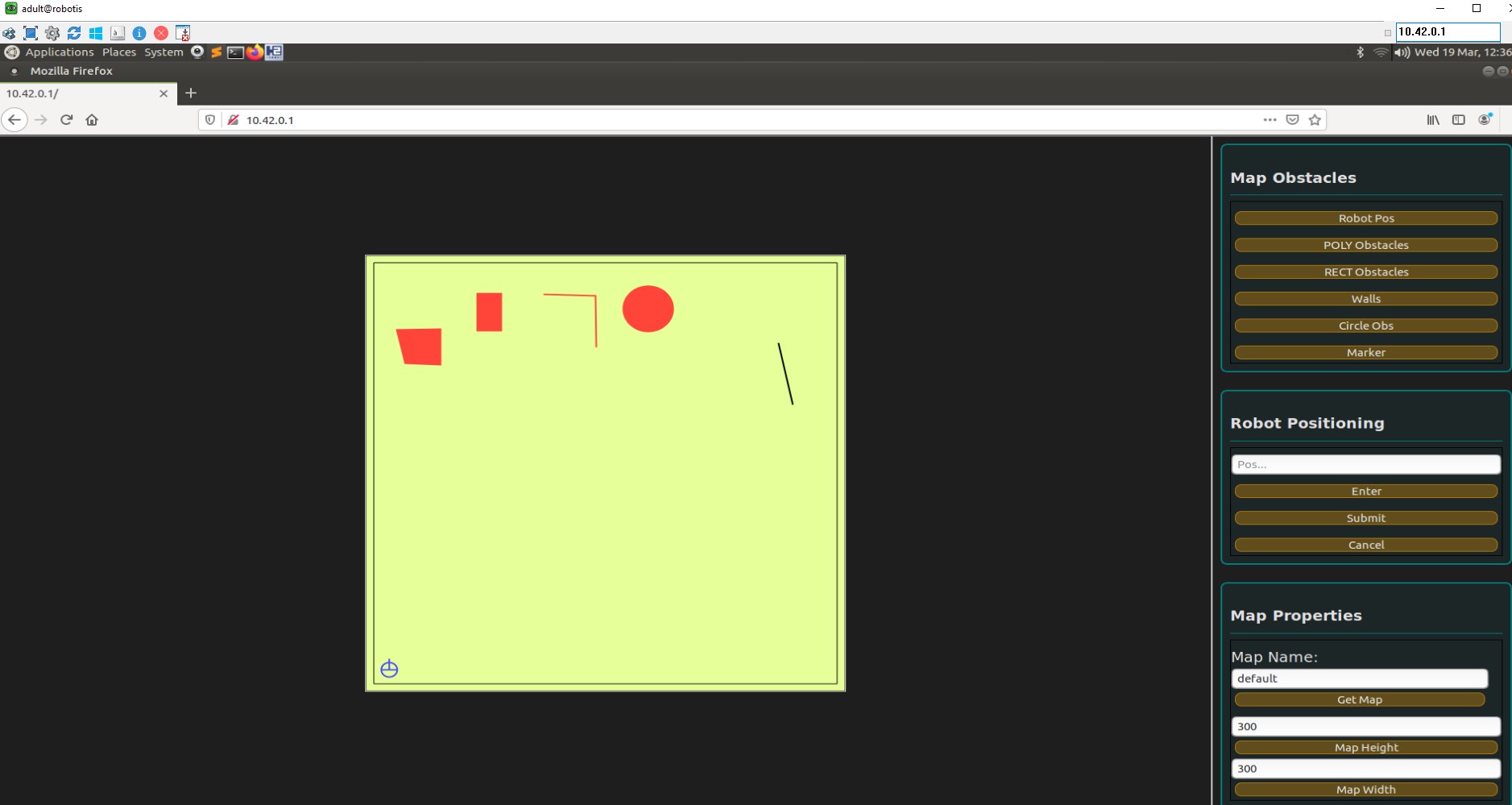}
    \caption{Image of new map layout with map on left and operation buttons on the right}
    \label{fig:new_map_image}
\end{figure}

Firstly, one of the main features that was present in the old system was the layout map as shown in Fig.~\ref{fig:old_map_image}. This map would be created and modified by the user and would depict a version of the obstacle cores from a bird-eye perspective. In this map, you would be able to modify the obstacles and the robot's position within it. When movements on the robot were completed, it would also depict the robot moving as the user entered commands. This feature was present on the same screen as the operation, and depending on the map size, it could vastly over-clutter the operation area. This feature, although good in practice, would clutter the screen and work only sometimes in the application. To make this feature more user-friendly, it was moved to the first screen where it has its fixed-size location on screen as shown in Fig.~\ref {fig:new_map_image}. As well as making it available on the second screen, should the user want to use it. While this optional second screen feature is not completed due to time constraints, the new layout of it being on the first screen reduces clutter and allows the user to focus either on creating the map or operating the robot, rather than having a feature you are not using being constantly on your screen. 

Another Major refinement between the old and the new versions of the UI is the dynamic page sizing as shown in Fig.~\ref{fig:new_small}. This feature will be analysed in the coming chapter, but just know that the new system provides a system for dynamic window sizing in a simple and user-friendly manner. Whereas the old system just compressed everything within the window, causing loss of page structure and readability. Seeing as shrinking this window in competition is a user-friendly to save processing power is common, this feature refinement will prove to be critical for the user experience.

The next general refinement of the code involved the ROS communication for the movement of the robot. The old system did not provide any actions such as left and right walks, even though the buttons were present, and the mapping of the configurations of these walks were not present, so updates to these walking configurations were added. Along with this, a restructuring of this message mapping code present in rosactions.js was refactored to be more readable and expandable in case more motions are added in the future. The buttons for the action buttons, such as crawling and recovery, were also implemented in a way such that developers in the future may be able to just add another button to the sub-menu and map it to its action in ROS without much hassle. This addition helps achieve the secondary goal of allowing the code to be scalable for future events.

On the final iteration of development, the addition of many of the buttons present on the right side being activated to turn into sub-menus provides users with less clutter, as shown in Fig.~\ref{fig:sub_menu}. This allows the operator to be more focused on the task at hand and reduce the cognitive load of performing the tasks, helping user performance in general applications \cite{Norman2013, Tidwell2010}. This also has the advantage of allowing more advanced users to keep tabs like the control list closed so they can focus on the task, and new users can open this legend of controls. We added the additional feature of tooltips to the buttons for the system, where almost every button has a descriptive tooltip that appears if operators hover over the button, as shown in Fig.~\ref{fig:movement_adjustment}. This visual labelling ensures that even non-expert users can immediately recognize the function without needing prior training. This aligns with best common practices and is in user-friendly development \cite{Cooper2014}.

The old system utilized a slider bar to represent the position in terms of how far its head is turned in a certain direction. This representation of the head's position proved to be valuable as the user can easily see if the head is aligned with the body without any guesswork involved. Tweaks to these sliders were made in terms of visual appeal and location on the screen with labelling to make it more clear to the user in terms of their use ~\ref{fig:sub_menu}. Additionally, the implementation of a reset button on the slider proved useful as it allowed the user to quickly reset the camera position without additional time consumed.

\subsection{Comparison setup}

The newly developed UI is evaluated by directly comparing it to the old system previously used on the robot. This comparison focuses on how each interface adheres to established UI development practices and maintains consistency. An interface that incorporates these practices more thoroughly is presumed to be more user-friendly. Although this approach does not offer definitive proof of improvement, it provides a qualitative measure to gauge which UI might better enhance the user experience.

\section{Results and Discussion}
This chapter reviews the completed teleoperation GUI, contrasts it with the legacy interface, and highlights key gains, remaining limits, and practical implications for competition use.

\subsection{Presentation of Results}

\begin{figure}[H]
  \centering
  \includegraphics[width=\textwidth]{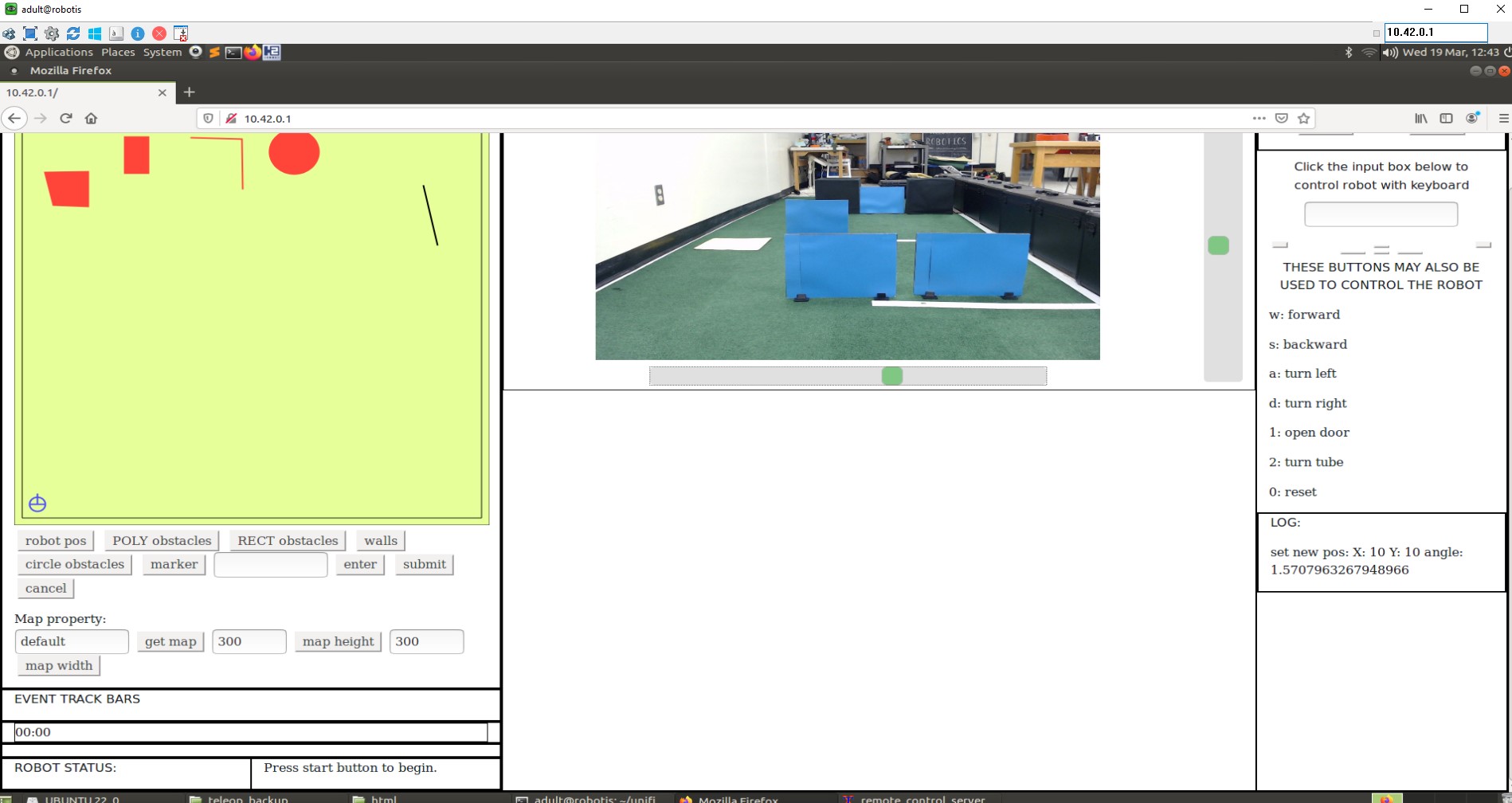}
  \caption{Legacy teleoperation screen.}
  \label{fig:old_UI_image}
\end{figure}

\begin{figure}[H]
  \centering
  \includegraphics[width=\textwidth]{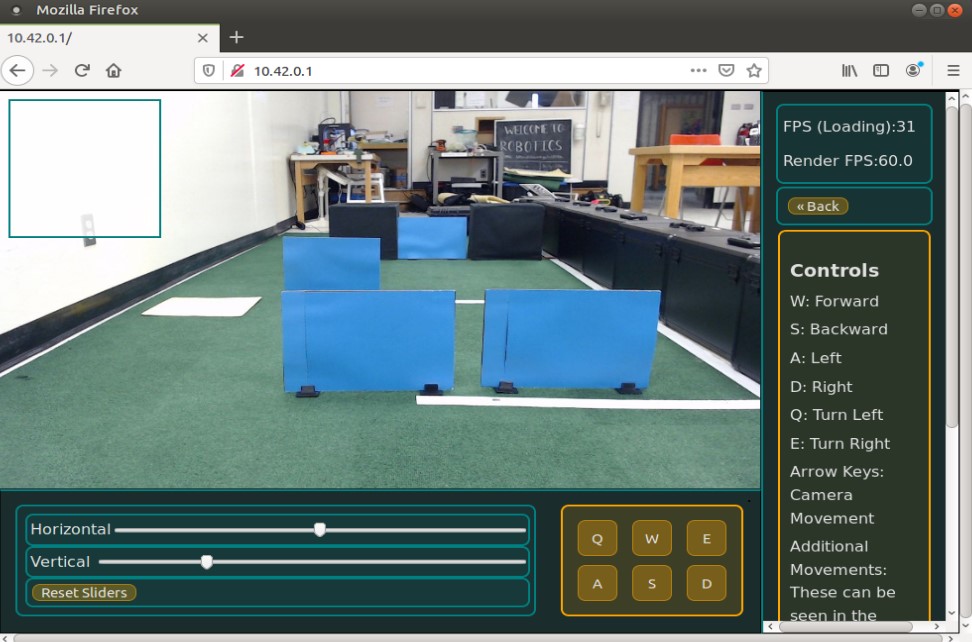}
  \caption{Sidebar with expanded sub-menu.}
  \label{fig:sub_menu}
\end{figure}

Figure~\ref{fig:old_UI_image} shows the former interface; Figure~\ref{fig:sub_menu} depicts the new one.

\paragraph{Layout \& Navigation}  
The new UI divides functions across two pages and clearly grouped panels, so related controls stay close and unused tools stay hidden. Separate “operate” and “map-edit” pages to remove clutter.

\paragraph{Colour Scheme}  
Dark blue backgrounds reduce glare, while bright-orange buttons stand out against both the background and live video, guiding the eye.

\paragraph{Camera Emphasis}  
The live feed is now central and larger.  Paired with keyboard shortcuts (WASD for motion, arrows for camera), operators can watch and act simultaneously.

\begin{figure}[H]
  \centering
  \includegraphics[width=0.5\textwidth]{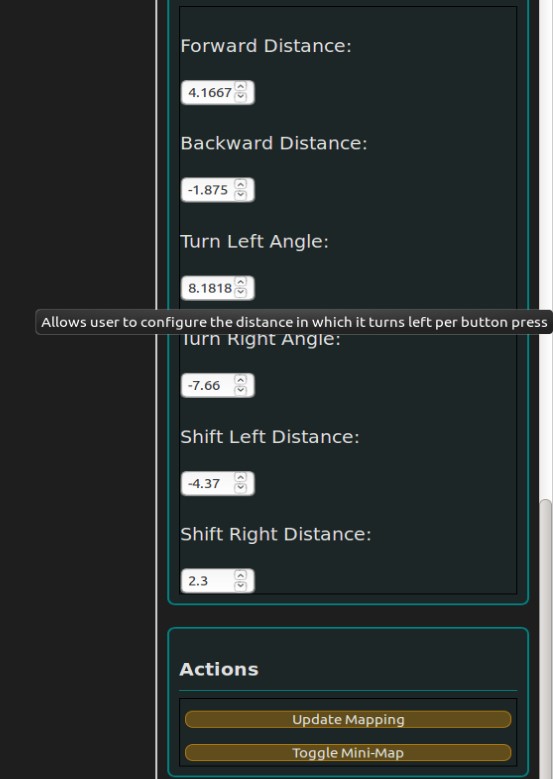}
  \caption{Movement-coefficient controls.}
  \label{fig:movement_adjustment}
\end{figure}

\paragraph{On-the-fly Tuning}  
Figure~\ref{fig:movement_adjustment} shows sliders for step length and turn angle; values persist locally so users adjust without restarting.

\begin{figure}[H]
  \centering
  \includegraphics[width=\textwidth]{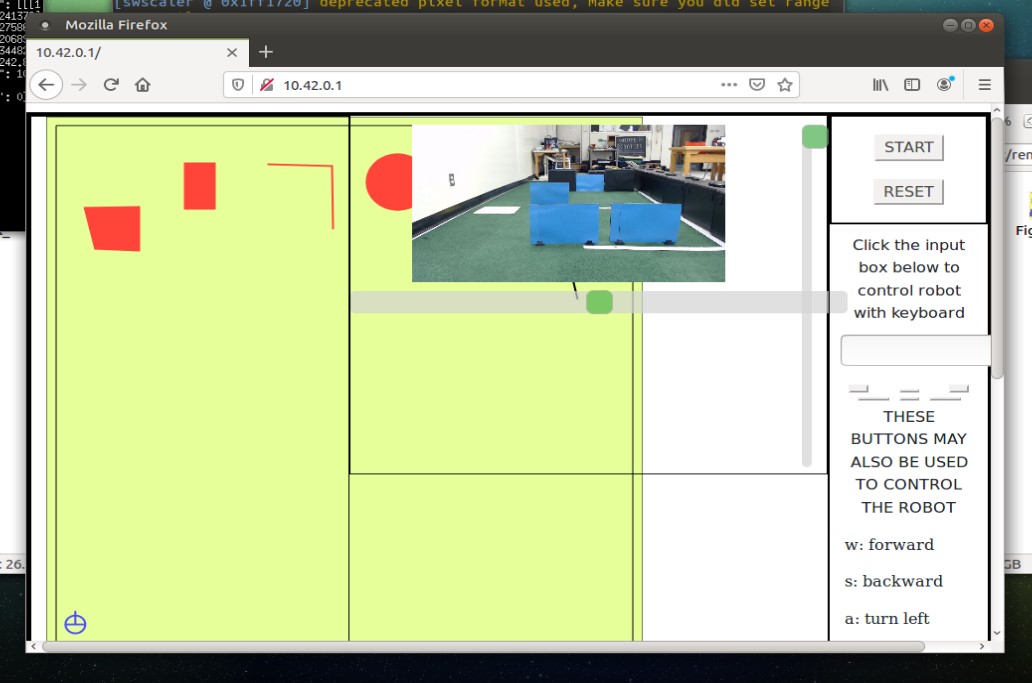}
  \caption{Old UI on a small display.}
  \label{fig:old_small}
\end{figure}

\begin{figure}[H]
  \centering
  \includegraphics[width=\textwidth]{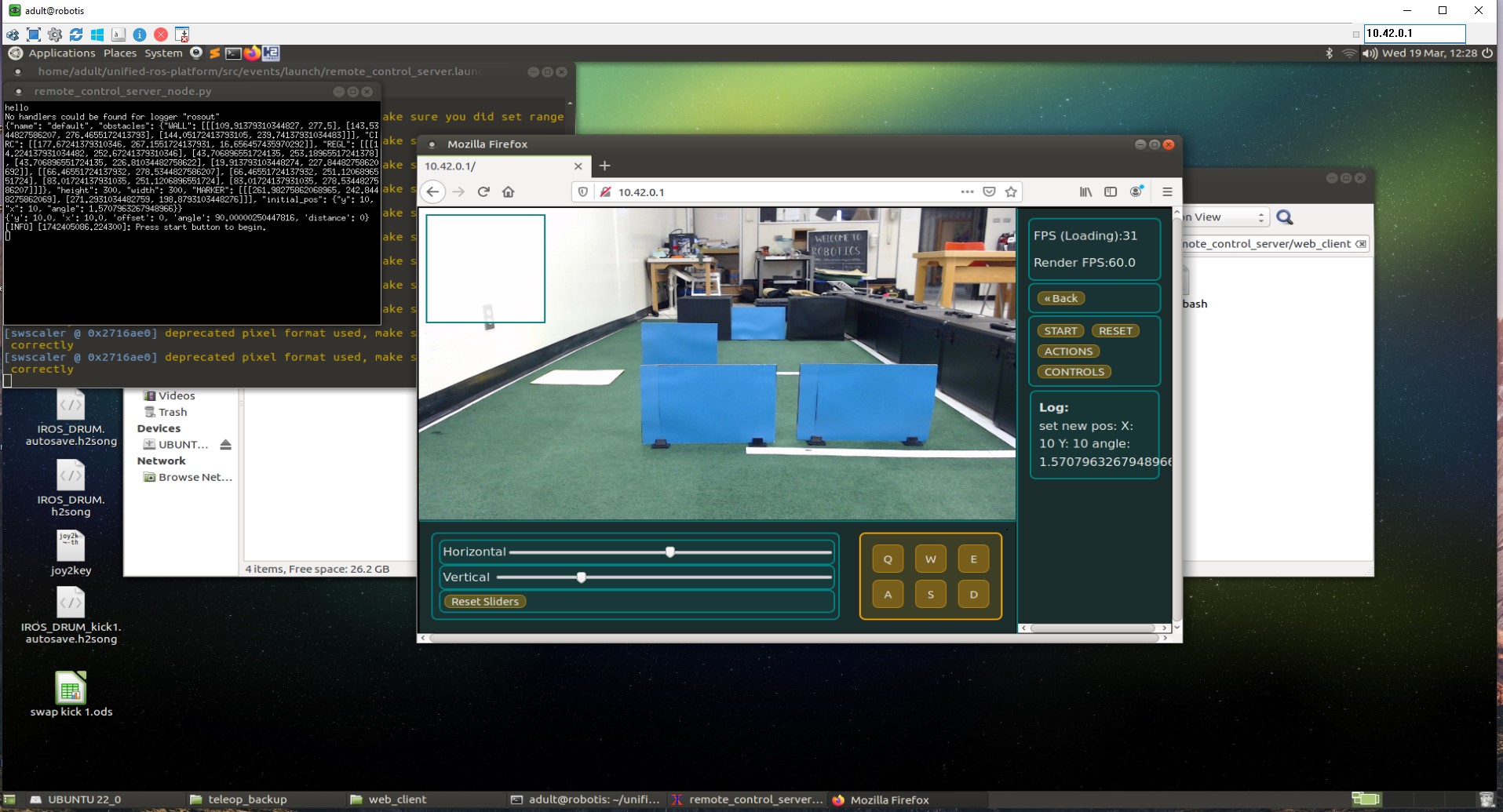}
  \caption{New UI on the same footprint.}
  \label{fig:new_small}
\end{figure}

\paragraph{Responsive Design}  
Flexbox scaling keeps panels legible on small windows (compare Figures~\ref{fig:old_small}–\ref{fig:new_small}). Scrollbars appear only when components outsize their areas.

\paragraph{Collapsible Menus}  
Sidebar buttons open drop-downs inside the same frame (Figure~\ref{fig:sub_menu}); each panel gets its own scrollbar to keep controls usable even in very narrow views.

\subsection{Discussion}

Notable advances include:

\begin{itemize}
  \item \textbf{Fine-tune sliders} for step/turn coefficients—absent before—let operators adapt mid-run.
  \item Added motions (crawl, self-stand) enable recovery and obstacle negotiation, impossible with the old tool.
  \item Relative to manual ROS commands or the legacy GUI, the new layout offers far less clutter, labelled controls, and reliable head-pan sliders, all reinforcing novice usability.
\end{itemize}

\subsection{Limitations}

Two key constraints remain: (i) minimal human-subject testing during iterations, limiting empirical validation of “user-friendly” claims; and (ii) tight timelines that postponed extra features that could further streamline competition performance.

\section{Conclusion}

The objective of this project was to create a user-friendly GUI designed for non-expert users to teleoperate with a humanoid robot through a FIRA-rated obstacle course.  This was done by addressing issues with the previous system used before this project. This newer version introduced significant improvements aimed at reducing cognitive load, increasing the general intuitiveness of the system, and ensuring that the current system is scalable for future endeavours.

The improvement of adding new motions and actions to complete the obstacle course proved an extreme success and was beneficial. Along with adding the bindings for those movements, they proved to be more important for the user experience with the screen UI than anticipated. Much of the time, it can be harder to navigate a screen if you are moving back and forth pressing different buttons, adding to the user's overstimulation. 

The addition of all the new layout features also leads to a very comfortable user navigation experience. The logically grouped controls and common placement of major controls (such as page navigation being in the top corners of the screen or confirmations being at the bottom of a section) provide a very simplistic navigation experience that should be familiar to most users in some way or another. Navigation of the basic application, as well as controlling the robot's mechanisms, is much easier because of this layout.

Although the application of this system is for a particular event on this robot, it was developed with the goal in mind to change how other UIs of a similar nature and tasks are developed and displayed. I think the improvement shown in this project could help others who are creating UIs to look deeper into the human-friendly layout of their systems, as it may provide more efficient operation and output from an individual user. This is an idea that is reinforced multiple times in similar literature, showing its underlying importance in development.

Although the absence of formal user testing and time limitations posed constraints on the scope of development, the resulting system represents a marked advancement over the old GUI for the event. Future research should focus on quantitative measurements and reviews with non-expert participants, in order to more directly develop the system.

Overall, this project has led to an improved system for the robot's competition-level event and may provide more results in the future with further development and study. It is important to develop the interactions that your user has with your system to be curated to the user, rather than the developer or the computer, as it may provide better results than one expects.

\subsection{Acknowledgement} 
We gratefully acknowledge the financial and institutional support provided by Laurentian University, particularly the Faculty of Science, Engineering and Architecture, and the School of Engineering and Computer Science. Their backing has been instrumental in the development and success of the Snobots team and the Laurentian Intelligent Mobile Robotics Lab (LIMRL).

We also extend our sincere thanks to our industrial sponsors, whose generous contributions have supported our operational and travel expenses, enabling us to participate in national and international competitions. Their support has been vital in advancing our research and outreach activities in the field of humanoid robotics.

\bibliographystyle{apsrev4-1}
\bibliography{references}  %%% Uncomment this line and comment out the ``thebibliography'' section below to use the external .bib file (using bibtex) .





% \end{thebibliography}

\end{document}